%% file: main.tex
\documentclass{llncs}
\usepackage{graphicx}

\usepackage{tikz}
\usepackage{comment}
\usepackage{amsmath,amssymb} 
\usepackage{color}

\usepackage{algorithmic}
\usepackage[ruled]{algorithm2e}
\usepackage{booktabs}
\usepackage{mathrsfs}
\usepackage{multirow}
\usepackage{color}
\usepackage{bbm}
\usepackage{bbding}
\usepackage{threeparttable}
\usepackage{xspace}
\usepackage{wrapfig}
\usepackage{setspace}
\usepackage{bm}

\makeatletter
\DeclareRobustCommand\onedot{\futurelet\@let@token\@onedot}
\def\@onedot{\ifx\@let@token.\else.\null\fi\xspace}

\def\ie{\emph{i.e}\onedot}
\def\eg{\emph{e.g}\onedot}
\def\etal{\emph{et al}\onedot}

\makeatother

\usepackage[pagebackref=true,breaklinks=true,colorlinks,bookmarks=false]{hyperref}

\begin{document}
\pagestyle{headings}
\mainmatter
\def\ECCVSubNumber{413}  

\title{On Mitigating Hard Clusters for \\Face Clustering}


\titlerunning{On Mitigating Hard Clusters for Face Clustering}
\author{Yingjie Chen\inst{1,2}\thanks{Equal contribution.} \and
Huasong Zhong\inst{2}$^{\star}$ \and Chong Chen\inst{2}\thanks{Corresponding author.} \and Chen Shen\inst{2} \and \\ Jianqiang Huang\inst{2} \and Tao Wang\inst{1} \and Yun Liang\inst{1} \and Qianru Sun\inst{3}}
\authorrunning{Y. Chen et al.}
%

\institute{
Peking University, Beijing, China
\email{\{chenyingjie,wangtao,ericlyun\}@pku.edu.cn}
\\
\and DAMO Academy, Alibaba Group, China
\email{cheung.cc@alibaba-inc.com,\{zjushenchen,zhonghsuestc,jianqiang.jqh\}@gmail.com}
\\
\and Singapore Management University, Singapore
\email{qianrusun@smu.edu.sg}
\\}

\maketitle

\input{sections/abstract}

\input{sections/introduction}
\input{sections/relatedwork}

\input{sections/method}

\input{sections/experiments}

\input{sections/conclusion}

~\\
\textbf{Acknowledgments.}
This work is supported by the National Key R\&D Program of China under Grant 2020AAA0103901, Alibaba Group through Alibaba Research Intern Program, and Alibaba Innovative Research (AIR) programme.

\clearpage
%
%
\bibliographystyle{splncs04}
\bibliography{egbib}

\end{document}

%% file: sections/abstract.tex
\begin{abstract}
Face clustering is a promising way to scale up face recognition systems using large-scale unlabeled face images. It remains challenging to identify small or sparse face image clusters that we call hard clusters, which is caused by the heterogeneity, \ie, high variations in size and sparsity, of the clusters. Consequently, the conventional way of using a uniform threshold (to identify clusters) often leads to a terrible misclassification for the samples that should belong to hard clusters. We tackle this problem by leveraging the neighborhood information of samples and inferring the cluster memberships (of samples) in a probabilistic way. We introduce two novel modules, Neighborhood-Diffusion-based Density (NDDe) and Transition-Probability-based Distance (TPDi), based on which we can simply apply the standard Density Peak Clustering algorithm with a uniform threshold. Our experiments on multiple benchmarks show that each module contributes to the final performance of our method, and by incorporating them into other advanced face clustering methods, these two modules can boost the performance of these methods to a new state-of-the-art.
Code is available at: \href{https://github.com/echoanran/On-Mitigating-Hard-Clusters}{https://github.com/echoanran/On-Mitigating-Hard-Clusters}.

%

\keywords{Face clustering \and Diffusion density \and Density peak clustering}
\end{abstract}

%% file: sections/introduction.tex
\section{Introduction}

Face recognition is a classical computer vision task~\cite{zhao2003face,parkhi2015deep,kortli2020face} that aims to infer person identities from face images.
Scaling it up relies on more annotated data if using deeper models. Face clustering is a popular and efficient solution to reducing the annotation costs~\cite{liu2016large,wang2018cosface,liu2017sphereface,deng2019arcface}.
%
%
%

\begin{figure}
    \centering
    \includegraphics[width=1.0\textwidth]{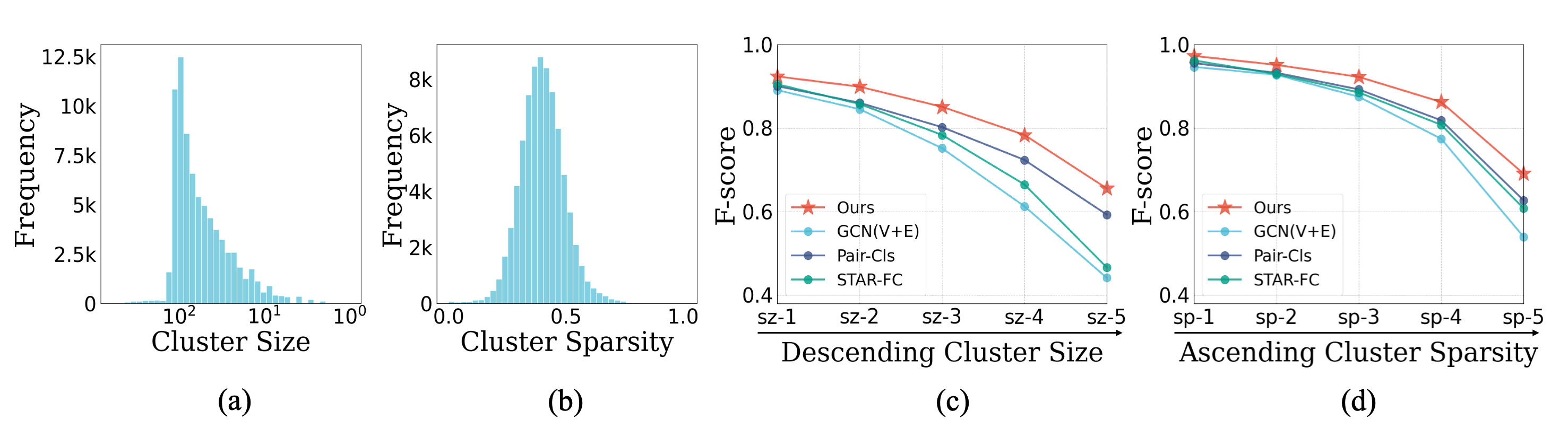}
    \caption{
    (a) and (b) show the ground-truth distribution of the face identity clusters on MS1M 5.21M dataset~\cite{ms1m}.
    (a) is for cluster size,~\ie, the number of samples in a cluster. (b) is for cluster sparsity, which is defined as the average cosine distance of all pair of samples in a cluster,~\eg, for cluster $C$ we have $\text{Sparsity}(C) = 1 - \frac{\sum_{ij\in C, i\neq j}\text{cosine}<x_{i},x_{j}>}{|C|(|C|-1)}$ where $|C|$ denotes the cluster size.
    %
    (c) and (d) show the performances (Pairwise F-score) of three top-performing methods, GCN(V+E)~\cite{yang2020learning}, Pair-Cls~\cite{liu2021learn}, and STAR-FC~\cite{shen2021structure}, compared to ours, on five cluster subsets with descending size (from sz-1 to sz-5) and ascending sparsity (from sp-1 to sp-5), respectively.}
    \label{fig:motivation}
\end{figure}



\begin{wrapfigure}[13]{r}{7cm}
    \centering
    \includegraphics[width=0.56\textwidth]{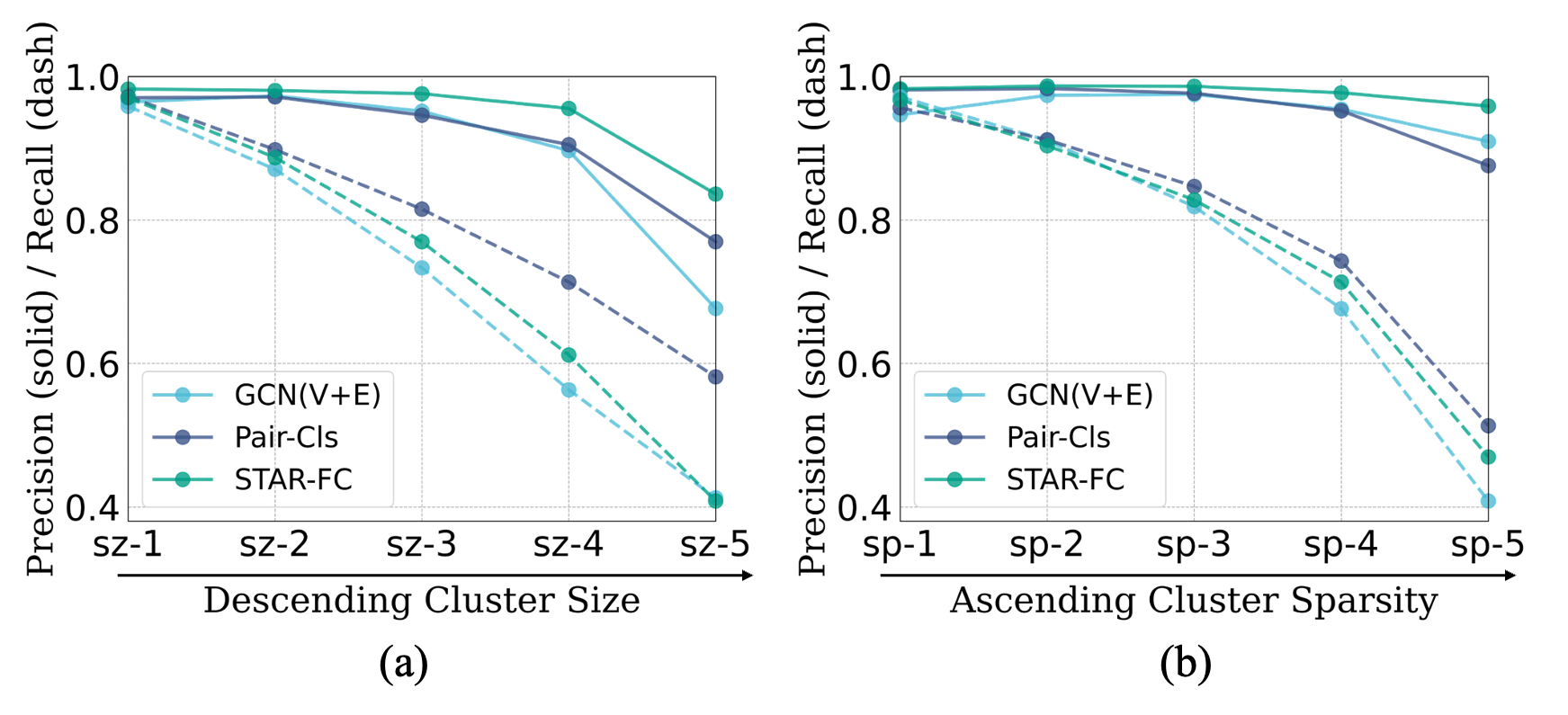}
    \caption{Pairwise precision and recall (of the three baselines) that elaborates the results in Figure~\ref{fig:motivation} (c) and (d). The recall of hard cluster subsets shows a significant drop.}
    \label{fig:motivation-2}
\end{wrapfigure}

\noindent
\textbf{Problems.}
Face clustering is challenging due to that 1) recognizing person identities is a fine-grained task; 2) the number of identities is always large, \eg, $77k$ on MS1M 5.21M dataset~\cite{ms1m}; and 3) the derived face clusters are often of high variations in both size and sparsity, and small or sparse clusters---we call \textbf{hard clusters}---are hard to identify. Figure~\ref{fig:motivation} (a) and (b) show the distributions of ground-truth clusters on MS1M 5.21M dataset.
%
For Figure~\ref{fig:motivation} (c) and (d), we first group these clusters into five subsets based on a fixed ranking of size and sparsity, respectively, and then evaluate three top-performing methods and ours on each subset.
It is clear that the performance drops significantly for hard clusters, \eg, in subsets sz-5 and sp-5, particularly on metric Recall (see Figure~\ref{fig:motivation-2}).
%
We think the reason is two-fold: 1) small clusters are overtaken by large ones; 2) samples of sparse clusters are wrongly taken as ``on'' low-density regions, \ie, the boundaries between dense clusters.

\setlength\intextsep{-4pt}

\begin{wrapfigure}[13]{r}{7.5cm}
    \centering
    \includegraphics[width=0.6\textwidth]{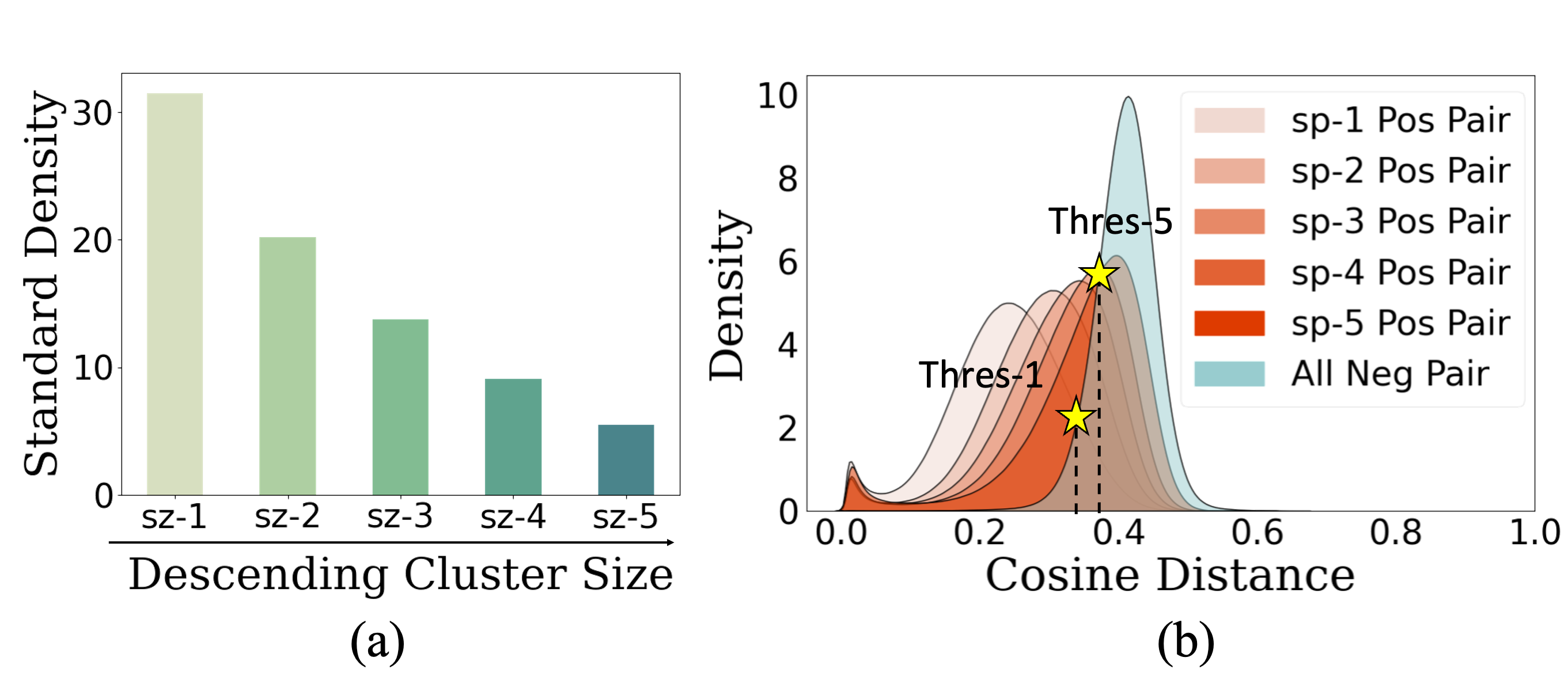}
    \caption{(a) The average standard density of clusters on each subset. (b) The probability density function on each subset with respect to the positive pairs. ``Pos'' indicates ``Positive'', and ``Neg'' for ``Negative''.}
    \label{fig:motivation_de_di}
\end{wrapfigure}

We elaborate these based on Density Peaking Clustering (DPC)~\cite{dpc} which has shown the impressive effectiveness in state-of-the-art face clustering works~\cite{yang2020learning,liu2021learn}.
DPC requires point-wise density and pair-wise distance to derive clustering results.
The density is usually defined as the number of neighbor points covered by an $\epsilon$-ball around each point~\cite{breiman1977variable}, and the distance is standard cosine distance. 
We find that both density and distance are highly influenced by the size and sparsity of latent clusters in face data.
%
For example, 1) smaller clusters tend to have lower density as shown in Figure~\ref{fig:motivation_de_di}~(a), so they could be misclassified as big ones by DPC, and 2) to identify positive pairs, higher-sparsity (lower-sparsity) clusters prefer a higher (lower) distance threshold, as indicated in Figure~\ref{fig:motivation_de_di}~(b), so it is hard to determine a uniform threshold for DPC.



\noindent
\textbf{Our Solution.}
Our clustering framework is based on DPC, and we aim to solve the above issues by introducing new definitions of point-wise density and pair-wise distance.
We propose a probabilistic method to derive a size-invariant density called Neighborhood-Diffusion-based Density (NDDe), and a sparsity-aware distance called Transition-Probability-based Distance (TPDi).
Applying DPC with NDDe and TPDi can mitigate hard clusters and yield efficient face clustering with a simple and uniform threshold.



\setlength\intextsep{-10pt}

We first build a transition matrix where each row contains the normalized similarities (predicted by a pre-trained model as in related works~\cite{wang2019linkage,yang2020learning,liu2021learn,shen2021structure}) between a point and its $K$-nearest neighbors, and each column is the transition probability vector from a point to the others.
%
Then, for NDDe, we specify a diffusion process on the matrix by 1)~initializing a uniform density for each point, and 2)~distributing the density to its $K$-nearest neighbors, where the distribution strength is proportional to the transition probability, until converge. 
The derived NDDe is invariant to the cluster size and thus free from the issue of small clusters. We provide the theoretical justification and empirical validation in Section~\ref{sec:NDDe}. 
For TPDi, we define a relative closeness that equals the inner product between two points' transition probability vectors (corresponding to two columns on the transition matrix).
%
We assume two points are close if they have similar transition probabilities to their common neighbors.
%
Our TPDi can yield more uniform sparsity (in clusters) than conventional distances such as cosine or Euclidean, and thus free from the issue of sparse clusters.
Our justification and validation are in Section~\ref{sec:TPDi}.

Our main contributions are threefold. 1) We inspect face clustering problem and find existing methods failed to identify hard clusters---yielding significantly low recall for small or sparse clusters. 
2) To mitigate the issue of small clusters, we introduce NDDe based on the diffusion of neighborhood densities.
3) To mitigate the issue of sparse clusters, we propose the relative distance TPDi that can facilitate a uniform sparsity in different clusters.
In experiments, we evaluate NDDe and TPDi on large-scale benchmarks and incorporate them into multiple baselines to show their efficiency.

%% file: sections/relatedwork.tex
\section{Related Work}
Face clustering has been extensively studied as an important task in the field of machine learning. Existing methods can be briefly divided into traditional methods and learning-based methods.
\paragraph{Traditional Methods.} 
Traditional methods include $K$-means~\cite{lloyd1982least}, HAC~\cite{sibson1973slink}, DBSCAN~\cite{ester1996density} and ARO~\cite{otto2017clustering}. These methods directly perform clustering on the extracted features without any supervision, and thus they usually seem simple but have obvious defects. $K$-means~\cite{lloyd1982least} assumes the cluster shape is convex and DBSCAN~\cite{ester1996density} assumes that the compactness of different clusters is homogeneous. The performances of these two methods are limited since both assumptions are impractical for face data.
The scale of unlabeled face images is usually large, and from this perspective, traditional methods are low-efficient and thus not suitable for face clustering task.
The computational efficiency of HAC~\cite{sibson1973slink} is not acceptable when handling millions of samples. 
To achieve better scalability, Otto~\etal~\cite{otto2017clustering} proposed ARO that uses an approximate rank-order similarity metric for clustering, but its performance is still far from satisfactory.

\paragraph{Learning-based Methods.} To improve the clustering performance, recent works \\ \cite{zhan2018consensus,wang2019linkage,yang2019learning,yang2020learning,guo2020density,shen2021structure,liu2021learn,wang2022ada} adopt a learning-based paradigm. Specifically, they first train a clustering model using a small part of data in a supervised manner and then test its performance on the rest of the data.
CDP~\cite{zhan2018consensus} proposed to aggregate the features extracted by different models, but the ensemble strategy results in a much higher computational cost.
L-GCN~\cite{wang2019linkage} first uses Graph Convolutional Networks (GCNs)~\cite{kipf2016semi} to predict the linkage in an instance pivot subgraph, and then extracts the connected components as clusters. LTC~\cite{yang2019learning} and GCN(V+E)~\cite{yang2020learning} both adopt two-stage GCNs for clustering with the whole $K$-NN graph. Specifically, LTC generates a series of subgraphs as proposals and detects face clusters thereon, and GCN(V+E) learns both confidence and connectivity via GCNs. To address the low-efficiency issue of GCNs, STAR-FC~\cite{shen2021structure} proposed a local graph learning strategy to simultaneously tackle the challenges of large-scale training and efficient inference. 
To address the noisy connections in the $K$-NN graph constructed in feature space, Ada-NETS~\cite{wang2022ada} proposed an adaptive neighbor discovery strategy to make clean graphs for GCNs. 
Although GCN-based methods have achieved significant improvements, they only use shallow GCNs resulting in a lack of high-order connection information, and in addition, their efficiency remains a problem. 
Pair-Cls~\cite{liu2021learn} proposed to use pairwise classification instead of GCNs to reduce memory consumption and inference time. Clusformer~\cite{nguyen2021clusformer} proposed an automatic visual clustering method based on Transformer~\cite{vaswani2017attention}. 

In general, existing learning-based methods have achieved significant improvements by focusing on developing deep models to learn better representation or pair-wise similarities, but they failed to identify and address the aforementioned hard cluster issues. In this paper, we explore face clustering task from a new perspective. Based on DPC~\cite{dpc}, we propose a size-invariant point-wise density NDDe and a sparsity-aware pair-wise distance TPDi, which can be incorporated into multiple existing methods for better clustering performance, especially on hard clusters.

%% file: sections/method.tex
\section{Preliminaries}
\label{sec:preliminaries}

\noindent
\textbf{Problem Formulation.}
Given $N$ unlabelled face images with numerical feature points $\bm{X} = \{x_{1}, x_{2}, \cdots, x_{N}\} \subset \bbbr^{N\times D}$, which are extracted by deep face recognition models, face clustering aims to separate these points into disjoint groups as $\bm{X} = \bm{X}_{1}\bigcup \bm{X}_{2}...\bigcup \bm{X}_{m}$, such that points with the same identity tend to be in the same group, while points with different identities tend to be in different groups.

\noindent
\textbf{Data Preprocessing.}
Following the general process of learning-based face clustering paradigm, the dataset $\bm{X}$ is split into a training set and a test set, $\bm{X} = \bm{X}_{\text{train}}\bigcup \bm{X}_{\text{test}}$. For a specific learning-based face clustering method, a clustering model is first trained on $\bm{X}_{\text{train}}$ in a supervised manner, and then the clustering performance is tested on $\bm{X}_{\text{test}}$. Without loss of generality, we always denote the features and labels as $\bm{X} = \{x_{1}, x_{2}, \cdots, x_{N}\}$ and $l = \{l_{1}, l_{2}, \cdots, l_{N}\}$, respectively, for both training stage and test stage.

\noindent
\textbf{Density Peak Clustering (DPC).}
DPC~\cite{dpc} identifies implicit cluster centers and assigns the remaining points to these clusters by connecting each of them to the higher density point nearby, which is adopted by several state-of-the-art face clustering methods~\cite{yang2020learning,liu2021learn}. In this paper, we also adopt DPC as the clustering algorithm. Given point-wise density $\rho = \{\rho_{1},\rho_{2},\cdots, \rho_{N}\}$ and pair-wise distance $(d_{ij})_{N\times N}$, for each point $i$, DPC first finds its nearest neighbor whose density is higher than itself, \ie,
$$\hat{j} = \text{argmin}_{\{j | \rho_{j} > \rho_{i}\}}d_{ij},$$
If $\hat{j}$ exists and $d_{i\hat{j}} < \tau$, then it connects $i$ to $\hat{j}$, where $\tau$ is a connecting threshold. In this way, these connected points form many separated trees, and each tree corresponds to a final cluster. Note that $\tau$ is uniform for all clusters, so consistent point-wise density $\rho$ and pair-wise distance $(d_{ij})_{N\times N}$ are essential for the success of DPC. To solve hard cluster issues, we propose a size-invariant density called Neighborhood-Diffusion-based Density (NDDe) and a sparsity-aware distance called Transition-Probability-based Distance (TPDi) for better $\rho$ and $(d_{ij})_{N\times N}$.

\section{Method}

\begin{figure*}[!t]
    \centering
    \includegraphics[width=0.99\textwidth]{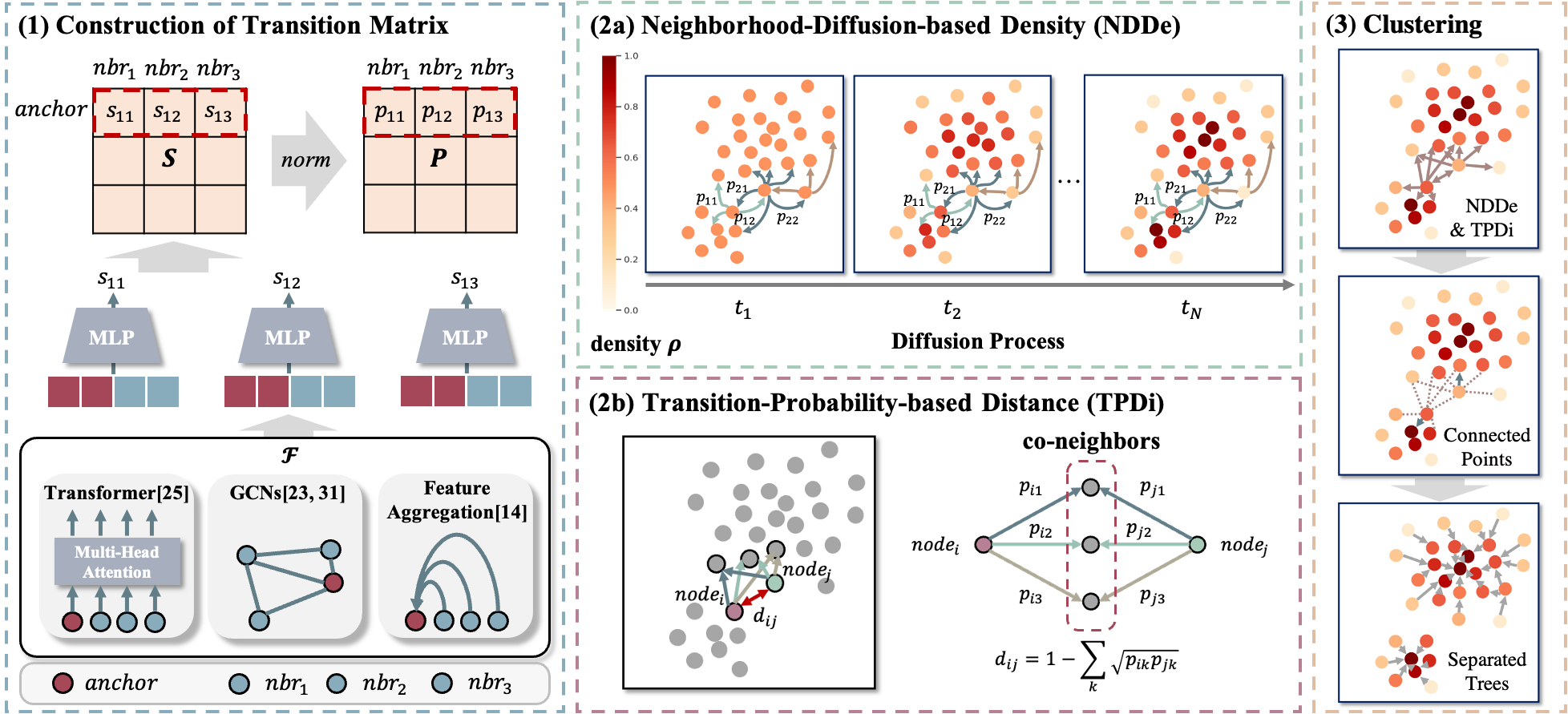}
    \caption{Overview. Our method consists of four steps: (1) Constructing transition matrix $\bm{P}$. Feature encoder $\mathcal{F}$ is for feature refinement, which can be Transformer, GCNs, or a feature aggregation module, and after that, an MLP is used to predict the similarity between each anchor point and its neighbors. (2a) Computing NDDe for each point through a diffusion process. (2b) Computing TPDi to measure the distance between points. (3) Applying DPC with NDDe and TPDi to obtain final clustering result.}
    \label{fig:overview}
\end{figure*}

Figure~\ref{fig:overview} shows the overall framework consisting of four steps.
First, we construct a transition matrix by learning the refined similarities between each point and its $K$-nearest neighbors using a model consisting of a feature encoder $\mathcal{F}$ and a Multi-Layer Perceptron (MLP).
%
The second step uses our first novel module: computing Neighborhood-Diffusion-based Density (NDDe) by diffusing point-wise density on the neighboring transition matrix, which is invariant to cluster size.
The third step is our second novel module: computing Transition-Probability-based Distance (TPDi) by introducing a relative closeness, which is aware of cluster sparsity. 
Fourth, we directly apply DPC with NDDe and TPDi to derive the final clustering result.


\subsection{Constructing Transition Matrix}
\label{sec:transition_matrix}
The standard way of construction the transition matrix is to compute the similarity between the deep features of pair-wise samples. 
%
The ``similarity'' can be the conventional cosine similarity or the learned similarity in more recent works such as~\cite{yang2020learning,liu2021learn,shen2021structure}. 
To reduce the memory consumption of using GCNs~\cite{yang2020learning,shen2021structure} for similarity learning, Pair-Cls~\cite{liu2021learn} simply learns the similarity via pair-wise classification by deciding whether two points share the same identity.
However, in Pair-Cls, all pairs are completely independent during training. We argue that the similarity between a point and one of its neighbors usually depends on the similarities between the point and its other neighbors.
Therefore, in our work, we adopt the same pair-wise classification, \ie, using an MLP to predict the similarity, and besides that, we leverage a collaborative prediction manner by considering the similarities between each point (as an anchor) and its neighbors as a whole to improve the robustness of the prediction, similar to~\cite{nguyen2021clusformer,liu2021learn}.

Here, we elaborate a general formulation. 
For a sample point $i$, we first find its $K$-nearest neighbors denoted as $\text{nbr}_{i} = \{i_{1},\cdots, i_{K}\}$, and then generate the following token sequence:
$$\tilde{x}_{i} = [x_{i}, x_{i_{1}}, x_{i_{2}}, \cdots, x_{i_{K}}].$$

Our similarity prediction model first takes $\tilde{x}_{i}$ as input, and outputs $K+1$ features after feature encoder $\mathcal{F}$:
$$\{t_{i}, t_{i_{1}}, \cdots, t_{i_{K}}\} = \mathcal{F}(\tilde{x}_{i}),$$
where $\mathcal{F}$ can be Transformer~\cite{vaswani2017attention}, GCNs~\cite{yang2020learning,shen2021structure} or a simple feature aggregation module~\cite{liu2021learn} (aggregate features of neighbors and concatenate to the feature of anchor). Then, for each neighbor $i_{j}$, $j = 1, ..., K$, $t_{i}$ are concatenated with $t_{i_{j}}$ and fed into an MLP with Sigmoid function to estimate the probability of that $i$ and $i_{j}$ share the same identity:
$$p_{ij} = \text{MLP}([t_{i}, t_{i_{j}}]).$$

Assuming $l_{ij}$ is the ground-truth label, $l_{ij}=1$ if $l_{i} = l_{i_{j}}$ and $l_{ij}=0$ vice versa. The total loss function is formulated as:
\begin{equation}
    \mathcal{L} = -\sum_{i=1}^{N}\sum_{j=1}^{K}(l_{ij}\log p_{ij} + (1-l_{ij})\log (1- p_{ij})).
\end{equation}

\noindent
Once the model converges, its predicted similarity takes the anchor's feature as well as its respective neighborhoods' features into consideration. 
Then, we can derive the similarity matrix $\bm{\hat{S}}_{N\times N}$ by applying this model on the test set. 

Finally, we assume $d_{i} = \sum_{j=1}^{N}\hat{s}_{ij}$ as the measure of the volume around point $i$, and generate the probability transition matrix $\bm{P}$ with each element as
$p_{ij} = {\hat{s}_{ij}}/{d_{i}}$.
The size of $\bm{P}$ is $N\times N$.
Please note that $\hat{S}$ is a sparse matrix where each row contains $K+1$ non-zero elements (itself and its top-$K$ nearest neighbors). Therefore, $\bm{P}$ is also sparse.
\textbf{We highlight} that the above approach is not the only way to construct the transition matrix $\bm{P}$, and we show the results of using other approaches to obtain $\bm{P}$ in the experiment section.

\subsection{Neighborhood-Diffusion-based Density}
\label{sec:NDDe}


In this section, we propose a new definition of the point-wise density, called NDDe, to alleviate the issue of small-size clusters.
In the transition matrix $\bm{P}$, each element $\bm{P}_{ij}$ denotes the probability from one point $i$ to its specific neighbor $j$.
%
It satisfies the conservation property, \ie, 
$\sum_{j}\bm{P}_{ij} = 1$, 
which induces a Markov chain on $\bm{X}$. 
Denoting $\bm{L} = \bm{I} - \bm{P}$ as the normalized graph Laplacian, where $\bm{I}$ is the identity matrix. We can specify a diffusion process as follows,
\begin{equation}
\label{eq:diffusion}
      \begin{cases}
    &\frac{\partial}{\partial t} \rho_{i}(t) = -\bm{L} \rho_{i}(t), \\
    & \rho_{i}(0) = 1.
     \end{cases}
\end{equation} 
where $\rho_{i}(t)$ is the density of point $i$ at $t$-th step.
Starting from a uniformly initialized density, the diffusion process keeps distributing the density of each point to its $K$-nearest neighbors, following the corresponding transition probabilities in $\bm{P}$,
until converged to a stationary distribution. 
The diffusion density thus can be induced as:
\begin{equation}
\label{eq:diffu-density}
    \rho_{i} = \lim_{t \to \infty} \rho_{i}(t).
\end{equation}

\paragraph{Justification of local properties of diffusion density.} 
The diffusion process is local because each point transits its density to $K$-nearest neighbors and itself (based on the transition matrix $\bm{P}$). 
If considering the ideal situation when $\bm{P}$ is closed, which means $p_{ij} > 0$ if and only if $x_{i}$ and $x_{j}$ share the same identity, we have the following theorem.
\begin{theorem}
\label{theorem:1}
Assume the dataset $\bm{X}$ can be split into $m$ disjoint clusters: \emph{\ie}, $\bm{X} = \bm{X}_{1}\bigcup...\bigcup \bm{X}_{m}$. Define $\bar{\rho}_{i} = \frac{\sum_{j\in \bm{X}_{i}}\rho(j)}{|\bm{X}_{i}|}$ is the average density of $\bm{X}_{i}$, and we have
$\bar{\rho}_{1} =\dots= \bar{\rho}_{m} = 1$
where $|\bm{X}_{i}|$ is the number of points in $\bm{X}_{i}$.
\end{theorem}

\begin{figure}[!t]
    \centering
    \includegraphics[width=1.0\textwidth]{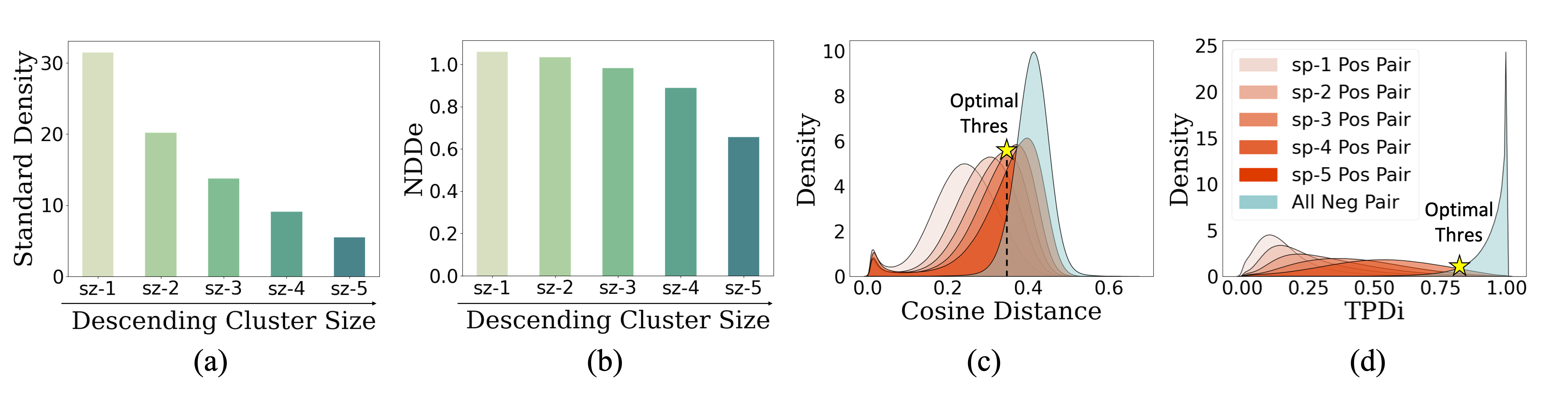}
    \setlength{\abovecaptionskip}{-5pt}
    \caption{(a) and (b) show the average values of the standard density and our NDDe on five cluster subsets from MS1M 5.21M dataset, respectively. NDDe is shown to be more uniform, i.e., small clusters are alleviated. (c) and (d) show the probability density functions of using the conventional cosine distance and TPDi on five cluster subsets from MS1M 5.21M dataset, respectively. Using TPDi makes it easier to decide a more uniform threshold to separate positive and negative pairs, in all subsets including the sparsest one ``sp-5''.}
    \label{fig:dis_den_comp}
\end{figure}

Theorem \ref{theorem:1} demonstrates that the average diffusion densities in all clusters are the same
regardless of cluster sizes.
In a dynamic sense, the diffusion process can elevate the density of latent small clusters, and thus enable DPC algorithm to identify density peaks in such clusters.
%
To further demonstrate our claim, we divide clusters in MS1M 5.21M dataset into five subsets according to cluster sizes and calculate the average diffusion density for each subset. As shown in Figure~\ref{fig:dis_den_comp}(a)(b), compared with the standard density, the average NDDe for different subsets are much more comparable.

\subsection{Transition-Probability-based Distance}
\label{sec:TPDi}



In this section, we introduces our new definition of the pair-wise distance, called TPDi, to solve the issue of varying sparsity in latent face clusters.
%
TPDi depicts the similarity between two points based on their respective transition probabilities (in $\bm{P}$) to the common neighbors. 
%
Assuming $C_{ij}=\text{nbr}_{i}\cap \text{nbr}_{j}$ contains the common neighbors in the $K$-nearest neighbors of both point $i$ and $j$. TPDi between them is defined as:
\newline
\begin{equation}
\label{eq:probability-distance}
    d_{ij} = 1 - \sum_{c\in C_{ij}}\sqrt{p_{ic}p_{jc}}.
\end{equation}

We highlight that TPDi has three impressive properties: (1) By Cauchy-Schwarz inequality, we have $\left(\sum_{c\in C_{ij}}\sqrt{p_{ic}p_{jc}}\right)^2\leq (\sum_{c\in C_{ij}}p_{ic})(\sum_{c\in C_{ij}}p_{jc})\leq 1$, so it is easy to check $0\leq d_{ij}\leq 1$, which implies that $d_{ij}$ can be a valid metric. (2) $d_{ij}= 0$ if and only if $p_{ic} = p_{jc}$ for all $c = 1,..., N$, which implies that $d_{ij}$ is small when $i$ and $j$ share as many as common neighbors. It is consistent with the motivation of TPDi. (3) Compared with cosine distance, TPDi of negative pairs and positive pairs are better separated, regardless of cluster sparsity (Figure \ref{fig:dis_den_comp}(c)(d)). So it is easier to choose a uniform threshold for TPDi.
\begin{remark}
If considering a simple case when each point transits to its neighbors with equal transition probability $\frac{1}{K}$, we have $d_{ij} = 1-\frac{2\text{Jaccard}(i,j)}{(1+\text{Jaccard}(i,j))}$, where $\text{Jaccard}(i,j)$ is the Jaccard similarity~\cite{ivchenko1998jaccard}. This implies that the TPDi is a generalization of Jaccard distance, which also demonstrate the feasibility of TPDi.
\end{remark}


\subsection{Overall Algorithm}
The overall clustering procedure is summarized in Algorithm~\ref{alg}. In our implementation, we use an iterative method as an approximation of Eq.~\ref{eq:diffu-density}.
\begin{algorithm}[!t]
\footnotesize
\setstretch{1.0}
\caption{Pseudocode for our method \label{alg}}
\LinesNumbered
\KwIn{Face dataset $\bm{X}=\{x_{1}, \dots, x_{N}\}$, number of nearest neighbors $K$, pre-trained similarity prediction model $\Phi$, convergence threshold $\epsilon$, connecting threshold $\tau$.}
\KwOut{clusters $\mathcal{C}$.}
\textbf{procedure} CLUSTERING \\
\qquad \textbf{for} each point $i$: \\
\qquad\qquad Find its $K$-nearest neighbors $\text{nbr}_{i}=\{i_k\}_{k=1}^K$ and construct $x_{i}^{\star}$; \\
\qquad\qquad Inference the similarities $\{s_{i,j}\}_{j=1}^K$ between them via $\Phi(x_{i}^{\star})$; \\
\qquad Obtain the pair-wise similarity matrix $\bm{\hat{S}}_{N\times N}$, and compute $\bm{P}_{N\times N}$; \\
\qquad Compute the point-wise density $\rho_{N\times 1}$ via NDDe($\bm{P}$); \\
\qquad Compute the pair-wise distance $(d_{ij})_{N\times N}$ via TPDi($\bm{P}$); \\
\qquad Obtain clusters $\mathcal{C}$ via DPC($\rho$, $(d_{ij})_{N\times N}$); \\
\textbf{end procedure} \\ 
\textbf{function} NDDe($\bm{P}$) \\
\qquad Initialize $\rho_{\text{pre}}=\{1\}_{N\times 1}$ \\
\qquad \textbf{while} $||\rho-\rho_{\text{pre}}||_{2} > \epsilon$: \\
\qquad\qquad $\rho=\rho_{\text{pre}}$;  $\rho_{\text{pre}}=\bm{P}\times \rho$;\\
\qquad \textbf{return} $\rho$ \\
\textbf{end function} \\
\textbf{function} TPDi($\bm{P}$) \\
\qquad \textbf{for} each pair of points $i$, $j$: \\
\qquad\qquad Compute $d_{ij}$ as shown in Eq.~\ref{eq:probability-distance}; \\
\quad\quad \textbf{return} $(d_{ij})_{N\times N}$ \\
\textbf{end function} \\
\end{algorithm}

%% file: sections/experiments.tex
\section{Experiments}

\subsection{Experimental Settings}
\subsubsection{Datasets.} 
We evaluate the proposed method on two public face clustering benchmark datasets, MS1M~\cite{ms1m} and  DeepFashion~\cite{deepfashion}. MS1M contains 5.8M images from 86K identities and the image representations are extracted by ArcFace~\cite{deng2019arcface}, which is a widely used face recognition model. MS1M is split into 10 almost equal parts officially. Following the same experimental protocol as in~\cite{yang2020learning,shen2021structure,liu2021learn}, we train our model on one labeled part and choose parts 1, 3, 5, 7, and 9 as unlabeled test data, resulting in five test subsets with sizes of 584K, 1.74M, 2.89M, 4.05M, and 5.21M images respectively. For DeepFashion dataset, following~\cite{yang2020learning}, we randomly sample 25,752 images from 3,997 categories for training and use the other 26,960 images with 3,984 categories for testing. 

\subsubsection{Metrics.} The performances of face clustering methods are evaluated using two commonly used clustering metrics, Pairwise F-score ($F_{P}$)~~\cite{pairwise} and BCubed F-score ($F_{B}$)~~\cite{bcubed}. Both metrics are reflections of precision and recall.

\subsubsection{Implementation Details.}
Our similarity prediction model consists of one transformer encoder layer~\cite{vaswani2017transformer} as $\mathcal{F}$ and an MLP. The input feature dimension, feedforward dimension, number of heads for $\mathcal{F}$ are set to 256, 2048, 8, respectively. LayerNorm~\cite{ba2016layer} is applied before Multi-head Attention module and Feed Forward module in $\mathcal{F}$, according to~\cite{xiong2020layer}. 
Dropout is set to 0.2. 
The MLP consists of three linear layers ($512\rightarrow 256, 256\rightarrow 128, 128\rightarrow 1$) with ReLU as the activation function for the first two layers and Sigmoid for the last layer.
Adam~\cite{kingma2014adam} is used for optimization.
For the computation of NDDe, we set the number of top nearest neighbors $K$ to 80 for MS1M and 10 for DeepFashion (the same as previous works~\cite{liu2021learn,yang2020learning}). Convergence
threshold $\epsilon$ is set to 0.05. Connecting threshold $\tau$ is searched within the range of $[0.5, 0.9]$ with a step of 0.05 on MS1M 584K dataset, and is fixed to 0.7 for all experiments. 

\subsection{Method Comparison}
We compare the proposed method with a series of clustering baselines, including both traditional methods and learning-based methods. Traditional methods include $K$-means~\cite{lloyd1982least}, HAC~\cite{sibson1973slink},DBSCAN~\cite{ester1996density}, and ARO~\cite{otto2017clustering}.
Learning-based methods include CDP~\cite{zhan2018consensus}, L-GCN~\cite{wang2019linkage}, LTC~\cite{yang2019learning}, GCN (V+E)~\cite{yang2020learning}, Clusformer~\cite{nguyen2021clusformer}, Pair-Cls~\cite{liu2021learn}, STAR-FC~\cite{shen2021structure}, and Ada-NETS~\cite{wang2022ada}. 
Since NDDe and TPDi can be incorporated into existing face clustering methods for better performance, we also incorporate them into GCN (V+E), Pair-Cls, and STAR-FC by using the three methods to obtain the transition matrix $\bm{P}$, which are denoted as GCN(V+E)++, Pair-Cls++, and STAR-FC++, respectively.

\begin{table}[!t]
\scriptsize
\centering
\caption{Comparison on MS1M when training with 0.5M labeled face images and testing on five test subsets with different numbers of unlabeled face images. $F_P$, $F_B$ are reported. GCN(V+E)++, Pair-Cls++ and STAR-FC++ denote incorporating NDDe and TPDi into the corresponding methods. The best results are highlighted with \textbf{bold}.}
\begin{tabular}{l|cc|cc|cc|cc|cc}
\toprule
\#Images & 
\multicolumn{2}{c|}{584K} & 
\multicolumn{2}{c|}{1.74M} & 
\multicolumn{2}{c|}{2.89M} & 
\multicolumn{2}{c|}{4.05M} &
\multicolumn{2}{c}{5.21M} \\ \cline{1-11}
Method / Metrics  & $F_{P}$ & $F_{B}$  & $F_{P}$ & $F_{B}$ &$F_{P}$ & $F_{B}$ &$F_{P}$ & $F_{B}$ &$F_{P}$ & $F_{B}$\\
\midrule
$K$-means~\cite{lloyd1982least} & 79.21 & 81.23 & 73.04 & 75.20 & 69.83 & 72.34 & 67.90 & 70.57 & 66.47 & 69.42 \\
HAC~\cite{sibson1973slink} & 70.63 & 70.46 & 54.40 & 69.53 & 11.08 & 68.62 & 1.40 & 67.69 & 0.37 & 66.96 \\
DBSCAN~\cite{ester1996density} & 67.93 & 67.17 & 63.41 & 66.53 & 52.50 & 66.26 & 45.24 & 44.87 & 44.94 & 44.74 \\
ARO~\cite{otto2017clustering} & 13.60 & 17.00 & 8.78 & 12.42 & 7.30 & 10.96 & 6.86 & 10.50 & 6.35 & 10.01 \\
CDP~\cite{zhan2018consensus} & 75.02 & 78.70 & 70.75 & 75.82 &
69.51 & 74.58 & 68.62 & 73.62 & 68.06 & 72.92 \\
L-GCN~\cite{wang2019linkage} & 78.68 & 84.37 & 75.83 & 81.61 & 74.29 & 80.11 & 73.70 & 79.33 & 72.99 & 78.60 \\
LTC~\cite{yang2019learning} & 85.66 & 85.52 & 82.41 & 83.01 & 80.32 & 81.10 & 78.98 & 79.84 & 77.87 & 78.86 \\
GCN(V+E)~\cite{yang2020learning} & 87.93 & 86.09 & 84.04 & 82.84 & 82.10 & 81.24 & 80.45 & 80.09 & 79.30 & 79.25 \\
Clusformer~\cite{nguyen2021clusformer} & 88.20 & 87.17 & 84.60 & 84.05 & 82.79 & 82.30 & 81.03 & 80.51 & 79.91 & 79.95 \\
Pair-Cls~\cite{liu2021learn} & 90.67 & 89.54 & 86.91 & 86.25 & 85.06 & 84.55 & 83.51 & 83.49 & 82.41 & 82.40 \\
STAR-FC~\cite{shen2021structure} & 91.97 & 90.21 & 88.28 & 86.26 & 86.17 & 84.13 & 84.70 & 82.63 & 83.46 & 81.47 \\
Ada-NETS~\cite{wang2022ada} & 92.79 & 91.40 & 89.33 & 87.98 & 87.50 & 86.03 & 85.40 & 84.48 & 83.99 & 83.28 \\
\midrule
GCN(V+E)++ 
& 90.72 & 89.28 & 86.06 & 84.36 & 85.97 & 84.24 & 84.76 & 83.10 & 83.69 & 82.26 \\
Pair-Cls++ 
& 91.70 & 89.94 & 88.17 & 86.50 & 86.49 & 84.76 & 85.25 & 83.50 & 83.74 & 82.61 \\
STAR-FC++ 
& 92.35 & 90.50 & 89.03 & 86.94 & 86.70 & 85.16 & 85.38 & 83.93 & 83.94 & 82.95 \\
\midrule
\textbf{Ours} &
\textbf{93.22} & \textbf{92.18} & \textbf{90.51} & \textbf{89.43} & \textbf{89.09} & \textbf{88.00} & \textbf{87.93} & \textbf{86.92} & \textbf{86.94} & \textbf{86.06} \\
\bottomrule
\end{tabular}
\label{tab:sota_ms1m}
\end{table}

\subsubsection{Results on MS1M.} 
Experimental results on MS1M dataset are shown in Table~\ref{tab:sota_ms1m}, which contains both $F_P$ and $F_B$ on five test subsets with different scales. We can observe that 1) Our method consistently outperforms the other methods in terms of both metrics, especially for large-scale subsets, \eg, the improvements of our method on 4.05M and 5.21M subsets are more than $2.5\%$. 2) By incorporating NDDe and TPDi into GCN (V+E), Pair-Cls and STAR-FC, their ++ versions achieve better clustering performance than the original versions, \eg, compared to GCN (V+E), the performance gains brought by GCN(V+E)++ are more than $3\%$ on large-scale test subsets, which demonstrates that NDDe and TPDi can raise the performance of other methods to a new state-of-the-art.

\subsubsection{Results on Hard Clusters.}

To demonstrate that our method is capable of tackling the issues of small clusters and sparse clusters, we conduct experiments by adding NDDe and TPDi one by one to our baseline model, \ie, the model with the same transition matrix but the density and distance computed in the standard way. As shown in the last three rows in Table~\ref{tab:comp_cluster_size} and Table~\ref{tab:comp_cluster_sparsity}, both NDDe and TPDi have raised the performance of the baseline model to a new level, especially on hard clusters.
\begin{figure}
    \centering
    \includegraphics[width=1.0\textwidth]{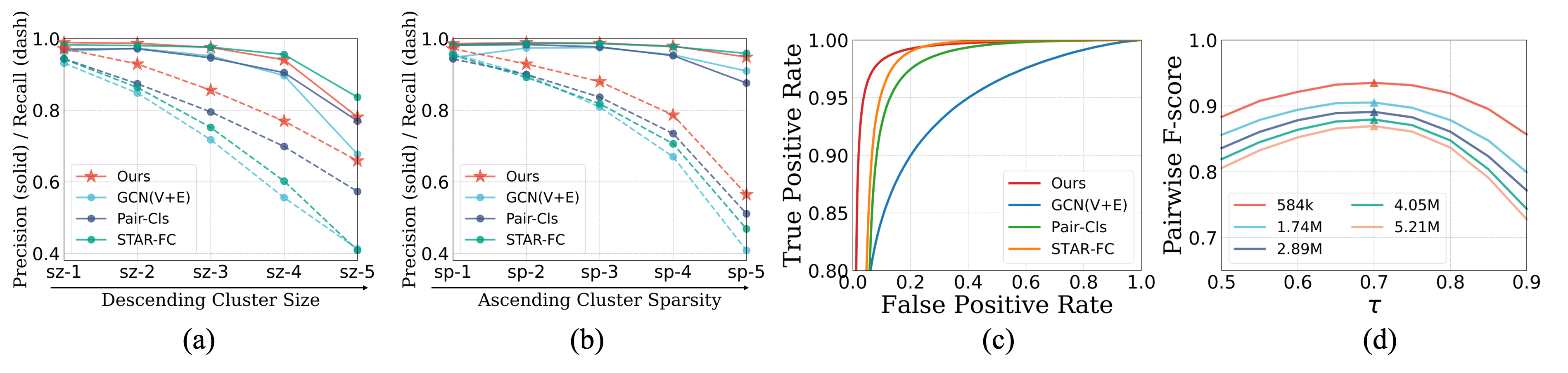}
    \setlength{\abovecaptionskip}{-5pt}
    \caption{(a) and (b) show Pairwise precision and recall of three baselines and our method. Significant improvements of our method in terms of recall can be observed. (c) ROC curves of the three baselines and our method. (d) Optimal threshold $\tau$ for the five test subsets of MS1M dataset.}
    \label{fig:comp_hyper}
\end{figure}
\begin{table}[!b]
\scriptsize
\centering
\caption{The effectiveness of NDDe and TPDi. $F_P$ and $F_B$ of five cluster subsets from MS1M 5.21M with descending size (from sz-1 to sz-5) are reported.}
\begin{tabular}{l|cc|cc|cc|cc|cc|cc}
\toprule
 &
\multicolumn{2}{c|}{sz-1} & 
\multicolumn{2}{c|}{sz-2} & 
\multicolumn{2}{c|}{sz-3} & 
\multicolumn{2}{c|}{sz-4} &
\multicolumn{2}{c|}{sz-5} & \multicolumn{2}{c}{total} \\ \cline{2-13}
& $F_{P}$ & $F_{B}$  & $F_{P}$ & $F_{B}$ &$F_{P}$ & $F_{B}$ &$F_{P}$ & $F_{B}$ &$F_{P}$ & $F_{B}$ &$F_{P}$ & $F_{B}$\\
\midrule
GCN(V+E) & 89.06 & 90.52 & 84.52 & 84.81 & 75.17 & 75.84 & 61.28 & 63.03 & 44.15 & 52.49 & 78.77 & 79.08  \\
Pair-Cls & 90.02 & 90.65 & 86.03 & 86.20 & 80.21 & 80.80 & 72.37 & 73.72 & 59.28 & 65.30 & 82.19 & 81.63 \\
STAR-FC & 90.47 & 91.13 & 85.75 & 86.11 & 78.35 & 78.78 & 66.49 & 67.44 & 46.65 & 51.21 & 83.74 & 82.00 \\
\midrule
Baseline & 57.54 & 63.45 & 52.39 & 55.89 & 43.84 & 47.62 & 37.84 & 41.77 & 34.67 & 42.23 & 41.49 & 50.76  \\
+NDDe & 83.67 & 86.06 & 78.38 & 78.95 & 69.63 & 70.28 & 60.19 & 61.49 & 49.85 & 54.53 & 72.47 & 74.39 \\
+TPDi(Ours) 
& 92.35 & 93.18 & 89.88 & 89.91 & 85.08 & 85.28 & 78.35 & 79.19 & 65.56 & 71.33 & 86.94 & 86.06 \\
\bottomrule
\end{tabular}
\label{tab:comp_cluster_size}
\end{table}
\begin{table}[!b]
\scriptsize
\centering
\caption{The effectiveness of NDDe and TPDi. $F_P$ and $F_B$ of five cluster subsets from MS1M 5.21M with ascending sparsity (from sp-1 to sp-5) are reported.}
\begin{tabular}{l|cc|cc|cc|cc|cc|cc}
\toprule
 & 
\multicolumn{2}{c|}{sp-1} & 
\multicolumn{2}{c|}{sp-2} & 
\multicolumn{2}{c|}{sp-3} & 
\multicolumn{2}{c|}{sp-4} &
\multicolumn{2}{c|}{sp-5} &
\multicolumn{2}{c}{total} \\ \cline{2-13}
& $F_{P}$ & $F_{B}$  & $F_{P}$ & $F_{B}$ &$F_{P}$ & $F_{B}$ &$F_{P}$ & $F_{B}$ &$F_{P}$ & $F_{B}$ &$F_{P}$ & $F_{B}$\\
\midrule
GCN(V+E) & 94.63 & 94.66 & 92.73 & 91.52 & 87.47 & 85.13 & 77.44 & 73.09 & 53.99 & 45.95 & 78.77 & 79.08  \\
Pair-Cls & 95.52 & 95.24 & 93.22 & 92.46 & 89.24 & 87.66 & 81.84 & 78.84 & 62.73 & 57.51 & 82.19 & 81.63  \\
STAR-FC & 96.18 & 95.27 & 92.92 & 91.50 & 88.50 & 85.96 & 80.78 & 76.54 & 60.81 & 53.56 & 83.74 & 82.00 \\
\midrule
Baseline & 63.16 & 63.84 & 62.23 & 62.95 & 57.32 & 58.09 & 49.19 & 50.20 & 32.98 & 35.48 & 41.49 & 50.76  \\
+NDDe & 92.30 & 91.00 & 87.47 & 85.70 & 82.11 & 79.30 & 72.69 & 69.97 & 52.53 & 51.17 & 72.47 & 74.39 \\
+TPDi(Ours) 
& 97.25 & 96.96 & 95.10 & 94.59 & 92.24 & 91.08 & 86.23 & 84.23 & 69.08 & 64.83 & 86.94 & 86.06 \\
\bottomrule
\end{tabular} \label{tab:comp_cluster_sparsity}
\end{table}

We also reproduce GCN(V+E), Pair-Cls and STAR-FC for comparison, all of which employ a clustering algorithm just as or similar to DPC, as shown in the first two rows in Table~\ref{tab:comp_cluster_size} and Table~\ref{tab:comp_cluster_sparsity}.
It is worth noticing that the improvements brought by our method over the three top-performing methods keep increasing on five cluster subsets with descending size or ascending sparsity. 
As shown in Figure~\ref{fig:comp_hyper}(a)(b), the improvements of our method in terms of Pairwise recall are more significant than Pairwise precision.
All the experimental results show the success of our method in mitigating hard clusters, owing to NDDe and TPDi.

\subsubsection{The Superiority of TPDi.} 
\begin{figure*}[!t]
    \centering
    \includegraphics[width=0.95\textwidth]{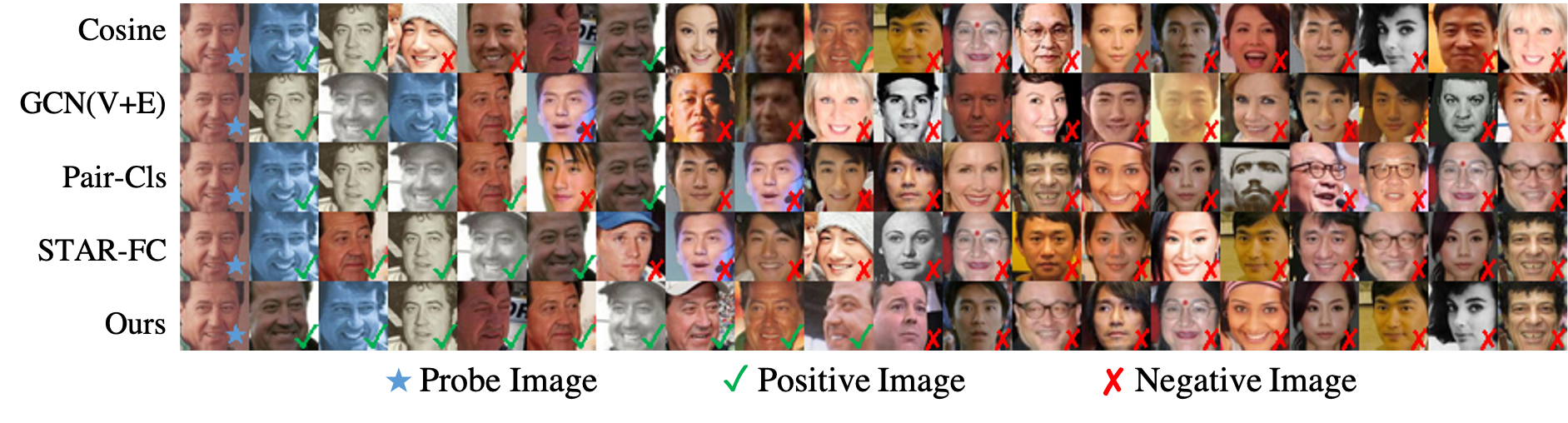}
    \setlength{\abovecaptionskip}{5pt}
    \caption{Top-20 images ranked by distance, using an image in hard clusters as probe.}
    \label{fig:case_study}
\end{figure*}

Figure~\ref{fig:comp_hyper}(c) shows the receiver operating characteristic (ROC) curves of three top-performing methods and ours, which are obtained by computing true/false positive rate at various distance threshold settings. Our method achieves the highest Area Under Curve (AUC), which illustrates that TPDi endows our method with a good measure of separability.
To show that by using TPDi, our method can yield efficient face clustering with a uniform connecting threshold $\tau$, we conduct experiments using different $\tau$ (from 0.5 to 0.9, with a step of 0.05) on all the test subsets of MS1M dataset, as shown in Figure \ref{fig:comp_hyper}(d). It can be observed that the best $\tau$ is the same for test subsets with varying scales. To be specific, given $\tau=0.7$, our method consistently achieves the highest $F_P$ on all test subsets.
Figure~\ref{fig:case_study} shows the discovery results of several methods with the image in the first column as a probe, and the images are ranked in ascending order of distance. We can observe that the discovery result of our method contains the most number of positive images.

\subsubsection{Results on DeepFashion.}

\setlength\intextsep{-10pt}
\begin{wraptable}[9]{r}{6.6cm}
\scriptsize
\centering
\caption{Comparison on DeepFashion. \#Clusters, $F_P$, $F_B$ and computing time are reported.}
\setlength{\tabcolsep}{1.3mm}{
\begin{tabular}{l|cccc}
\toprule
Method & \#Clusters & $F_{P}$ & $F_{B}$ & Time \\
\midrule
$K$-means~\cite{lloyd1982least} & 3991 & 32.86 & 53.77 & 573s \\
HAC~\cite{sibson1973slink} & 17410 & 22.54 & 48.7 & 112s \\
DBSCAN~\cite{ester1996density} & 14350 & 25.07 &53.23 & 2.2s \\
ARO~\cite{otto2017clustering} & 10504 & 26.03 & 53.01 & 6.7s \\
CDP~\cite{zhan2018consensus} & 6622 & 28.28 & 57.83 & 1.3s \\
L-GCN~\cite{wang2019linkage} & 10137 & 28.85 & 58.91 & 23.3s \\
LTC~\cite{yang2019learning} & 9246 & 29.14 & 59.11 & 13.1s \\
GCN(V+E)~\cite{yang2020learning} & 6079 & 38.47 & 60.06 & 18.5s \\
Pair-Cls~\cite{liu2021learn} & 6018 & 37.67 & 62.17 & 0.6s \\
STAR-FC~\cite{shen2021structure} & - & 37.07 & 60.60 & - \\
Ada-NETS~\cite{wang2022ada} & - & 39.30 & 61.05 & - \\
\midrule
\textbf{Ours} 
& 8484 & 40.91 & 63.61 & 4.2s \\
\bottomrule
\end{tabular}}
\label{tab:sota_deepfashion}
\end{wraptable}

For DeepFashion dataset, clustering task is much harder since it is an open set problem. It can be observed that our method also uniformly outperforms the other methods in terms of both $F_P$ and $F_B$ with comparable computing time, as shown in Table~\ref{tab:sota_deepfashion}.

\subsection{Ablation Study}
To demonstrate the effectiveness of NDDe and TPDi, we conduct an ablation study on MS1M 5.21M dataset, as shown in Table~\ref{tab:ablation}. All these four methods use the same transition matrix as described in Section~\ref{sec:transition_matrix}. $M_1$ is our baseline model, which uses the standard density and cosine distance. $M_2$ is obtained by replacing the cosine distance in $M_1$ with TPDi, $M_3$ is obtained by replacing the standard density in $M_1$ with NDDe, and $M_4$ is the proposed method using both NDDe and TPDi as the density $\rho$ and distance $(d_{ij})_{N\times N}$ required by DPC. 
Table~\ref{tab:ablation} shows that both NDDe and TPDi contribute to the final clustering performance. And the improvement brought by NDDe is more significant, which illustrates that NDDe is essential for the success of our method.

\begin{table}[!t]
\scriptsize
\centering
\caption{Ablation study of NDDe and TPDi on MS1M. $F_P$ and $F_B$ are reported.}
\begin{tabular}{l|cc|cc|cc|cc|cc|cc}
\toprule
& \multirow{2}{*}{NDDe} & \multirow{2}{*}{TPDi} &
\multicolumn{2}{c|}{584K} & 
\multicolumn{2}{c|}{1.74M} & 
\multicolumn{2}{c|}{2.89M} & 
\multicolumn{2}{c|}{4.05M} &
\multicolumn{2}{c}{5.21M} \\ \cline{4-13}
&  &  & $F_{P}$ & $F_{B}$  & $F_{P}$ & $F_{B}$ &$F_{P}$ & $F_{B}$ &$F_{P}$ & $F_{B}$ &$F_{P}$ & $F_{B}$\\
\midrule
$M_{1}$ &   &   &
53.03 & 56.75 & 47.80 & 53.84 & 45.07 & 52.41 & 43.29 & 51.56 & 41.49 & 50.76 \\
$M_{2}$ &   & \Checkmark & 
61.07 & 59.81 & 59.29 & 58.26 & 58.66 & 57.40 & 58.37 & 57.00 & 57.88 & 56.48 \\
$M_{3}$ & \Checkmark &   &  
82.98 & 80.33 & 78.79 & 77.87 & 76.32 & 76.42 & 74.08 & 75.28 & 72.47 & 74.39 \\
$M_{4}$ & \Checkmark & \Checkmark &
\textbf{93.22} & \textbf{92.18} & \textbf{90.51} & \textbf{89.43} & \textbf{89.09} & \textbf{88.00} & \textbf{87.93} & \textbf{86.92} & \textbf{86.94} & \textbf{86.06} \\
\bottomrule
\end{tabular}
\label{tab:ablation}
\end{table}

\setlength\intextsep{-10pt}
\begin{wrapfigure}[15]{r}{6.5cm}
    \centering
    \includegraphics[width=0.5\textwidth]{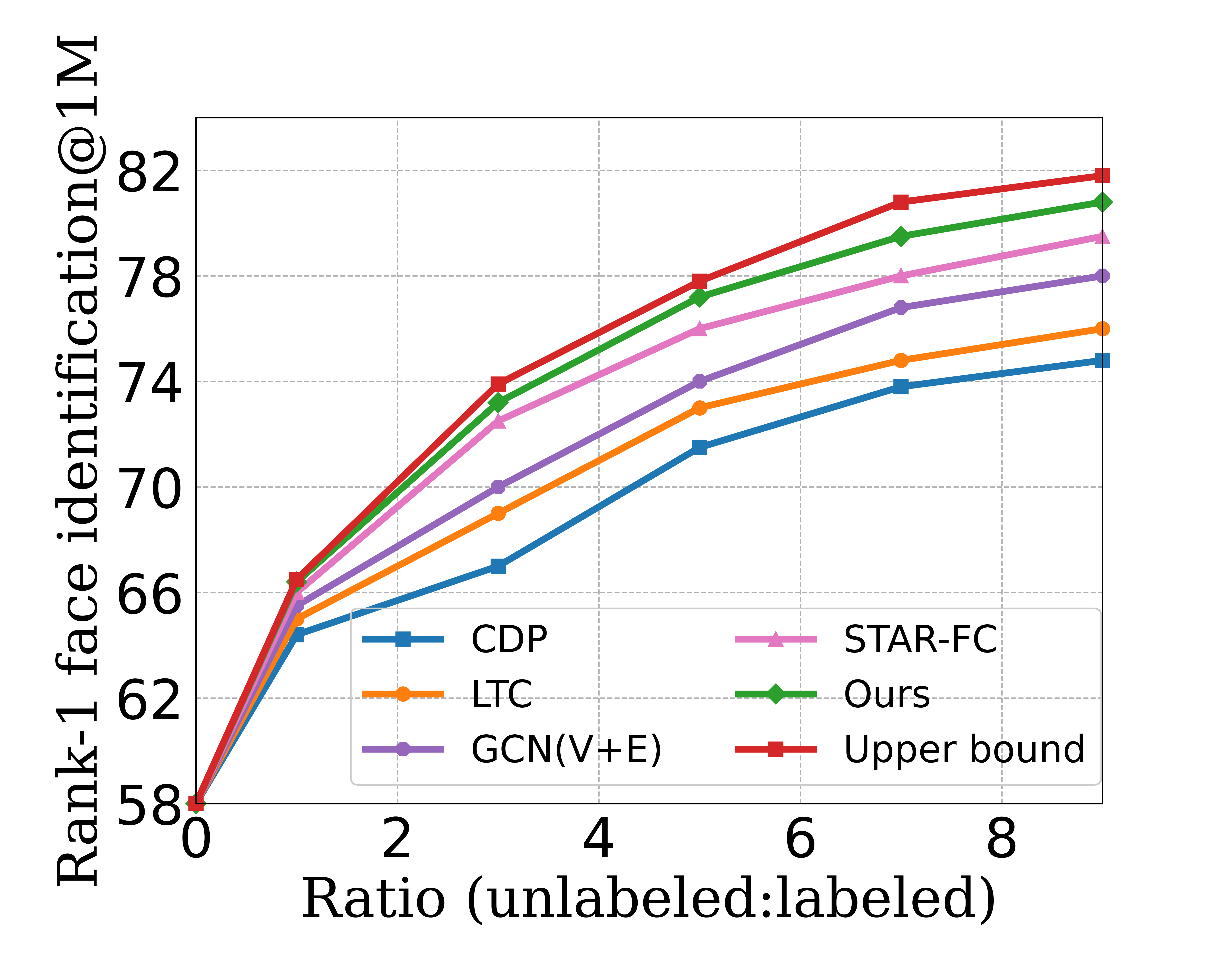}
    \setlength{\abovecaptionskip}{5pt}
    \caption{Rank-1 face identification accuracy on MegaFace with 1M distractors.}
    \label{fig:recognition}
\end{wrapfigure}
\subsection{Face Recognition}
To further show the potential of our method in scaling up face recognition systems using large-scale unlabeled face images, we use our method to generate pseudo-labels for unlabeled face images and use them to train face recognition models.
For a fair comparison, we adopt the same experimental setting as in~\cite{shen2021structure,yang2019learning,yang2020learning}. 
We use a fixed number of labeled data and different ratios of unlabeled data with pseudo-labels 
to train face recognition models and test their performance on MegaFace benchmark~\cite{kemelmacher2016megaface} taking the rank-1 face identification accuracy with 1M distractors as metric.
In Figure~\ref{fig:recognition}, the upper bound is trained by assuming all unlabeled data have ground-truth labels, and the other five curves illustrate that all the methods benefit from an increase of the unlabeled data with pseudo-labels.
And it can be observed that our method consistently achieves the highest performance given any ratio of unlabeled data, and improves the performance of the face recognition model from $58.20\%$ to $80.80\%$, which is the closest to the upper bound.

%% file: sections/conclusion.tex
\section{Conclusion}
In this paper, we point out a key issue in face clustering task---the low recall of hard clusters, \ie, small clusters and sparse clusters.
We find the reasons behind this are 1) smaller clusters tend to have a lower density, and 2) it is hard to set a uniform (distance) threshold to identify the clusters of varying sparsity.
We tackle the problems by proposing two novel modules, NDDe and TPDi, which yield the size-invariant density and the sparsity-aware distance, respectively.
Our extensive ablation study shows that each of them contributes to improving the recall on hard clusters, consistently on multiple face clustering benchmarks.

%% file: main.bbl
\begin{thebibliography}{10}
\providecommand{\url}[1]{\texttt{#1}}
\providecommand{\urlprefix}{URL }
\providecommand{\doi}[1]{https://doi.org/#1}

\bibitem{bcubed}
Amig{\'o}, E., Gonzalo, J., Artiles, J., Verdejo, F.: A comparison of extrinsic
  clustering evaluation metrics based on formal constraints. Information
  Retrieval  \textbf{12}(4),  461--486 (2009)

\bibitem{ba2016layer}
Ba, J.L., Kiros, J.R., Hinton, G.E.: Layer normalization. arXiv preprint
  arXiv:1607.06450  (2016)

\bibitem{pairwise}
Banerjee, A., Krumpelman, C., Ghosh, J., Basu, S., Mooney, R.J.: Model-based
  overlapping clustering. In: Proceedings of the eleventh ACM SIGKDD
  International Conference on Knowledge Discovery in Data Mining. pp. 532--537
  (2005)

\bibitem{breiman1977variable}
Breiman, L., Meisel, W., Purcell, E.: Variable kernel estimates of multivariate
  densities. Technometrics  \textbf{19}(2),  135--144 (1977)

\bibitem{deng2019arcface}
Deng, J., Guo, J., Xue, N., Zafeiriou, S.: Arc{F}ace: Additive angular margin
  loss for deep face recognition. In: CVPR (2019)

\bibitem{ester1996density}
Ester, M., Kriegel, H.P., Sander, J., Xu, X., et~al.: A density-based algorithm
  for discovering clusters in large spatial databases with noise. In: SIGKDD
  (1996)

\bibitem{guo2020density}
Guo, S., Xu, J., Chen, D., Zhang, C., Wang, X., Zhao, R.: Density-aware feature
  embedding for face clustering. In: CVPR (2020)

\bibitem{ms1m}
Guo, Y., Zhang, L., Hu, Y., He, X., Gao, J.: Ms-celeb-1m: A dataset and
  benchmark for large-scale face recognition. In: European Conference on
  Computer Vision. pp. 87--102. Springer (2016)

\bibitem{ivchenko1998jaccard}
Ivchenko, G., Honov, S.: On the jaccard similarity test. Journal of
  Mathematical Sciences  \textbf{88}(6),  789--794 (1998)

\bibitem{kemelmacher2016megaface}
Kemelmacher-Shlizerman, I., Seitz, S.M., Miller, D., Brossard, E.: The
  {M}ega{F}ace {B}enchmark: 1 million faces for recognition at scale. In: CVPR
  (2016)

\bibitem{kingma2014adam}
Kingma, D.P., Ba, J.: Adam: A method for stochastic optimization. arXiv
  preprint arXiv:1412.6980  (2014)

\bibitem{kipf2016semi}
Kipf, T.N., Welling, M.: Semi-supervised classification with graph
  convolutional networks. arXiv preprint arXiv:1609.02907  (2016)

\bibitem{kortli2020face}
Kortli, Y., Jridi, M., Al~Falou, A., Atri, M.: Face recognition systems: A
  survey. Sensors  \textbf{20}(2), ~342 (2020)

\bibitem{liu2021learn}
Liu, J., Qiu, D., Yan, P., Wei, X.: Learn to cluster faces via pairwise
  classification. In: Proceedings of the IEEE/CVF International Conference on
  Computer Vision. pp. 3845--3853 (2021)

\bibitem{liu2017sphereface}
Liu, W., Wen, Y., Yu, Z., Li, M., Raj, B., Song, L.: Sphereface: Deep
  hypersphere embedding for face recognition. In: CVPR (2017)

\bibitem{liu2016large}
Liu, W., Wen, Y., Yu, Z., Yang, M.: Large-margin softmax loss for convolutional
  neural networks. In: ICML (2016)

\bibitem{deepfashion}
Liu, Z., Luo, P., Qiu, S., Wang, X., Tang, X.: Deepfashion: Powering robust
  clothes recognition and retrieval with rich annotations. In: Proceedings of
  the IEEE conference on computer vision and pattern recognition. pp.
  1096--1104 (2016)

\bibitem{lloyd1982least}
Lloyd, S.: Least squares quantization in pcm. TIP  (1982)

\bibitem{nguyen2021clusformer}
Nguyen, X.B., Bui, D.T., Duong, C.N., Bui, T.D., Luu, K.: Clusformer: A
  transformer based clustering approach to unsupervised large-scale face and
  visual landmark recognition. In: Proceedings of the IEEE/CVF Conference on
  Computer Vision and Pattern Recognition. pp. 10847--10856 (2021)

\bibitem{otto2017clustering}
Otto, C., Wang, D., Jain, A.K.: Clustering millions of faces by identity. TPAMI
   (2017)

\bibitem{parkhi2015deep}
Parkhi, O.M., Vedaldi, A., Zisserman, A.: Deep face recognition  (2015)

\bibitem{dpc}
Rodriguez, A., Laio, A.: Clustering by fast search and find of density peaks.
  Science  \textbf{344}(6191),  1492--1496 (2014)

\bibitem{shen2021structure}
Shen, S., Li, W., Zhu, Z., Huang, G., Du, D., Lu, J., Zhou, J.: Structure-aware
  face clustering on a large-scale graph with 107 nodes. In: Proceedings of the
  IEEE/CVF Conference on Computer Vision and Pattern Recognition. pp.
  9085--9094 (2021)

\bibitem{sibson1973slink}
Sibson, R.: Slink: an optimally efficient algorithm for the single-link cluster
  method. The Computer Journal  (1973)

\bibitem{vaswani2017attention}
Vaswani, A., Shazeer, N., Parmar, N., Uszkoreit, J., Jones, L., Gomez, A.N.,
  Kaiser, {\L}., Polosukhin, I.: Attention is all you need. Advances in neural
  information processing systems  \textbf{30} (2017)

\bibitem{vaswani2017transformer}
Vaswani, A., Shazeer, N., Parmar, N., Uszkoreit, J., Jones, L., Gomez, A.N.,
  Kaiser, L., Polosukhin, I.: Attention is all you need. In: NIPS (2017)

\bibitem{wang2018cosface}
Wang, H., Wang, Y., Zhou, Z., Ji, X., Gong, D., Zhou, J., Li, Z., Liu, W.:
  Cosface: Large margin cosine loss for deep face recognition. In: CVPR (2018)

\bibitem{wang2022ada}
Wang, Y., Zhang, Y., Zhang, F., Wang, S., Lin, M., Zhang, Y., Sun, X.:
  Ada-nets: Face clustering via adaptive neighbour discovery in the structure
  space. arXiv preprint arXiv:2202.03800  (2022)

\bibitem{wang2019linkage}
Wang, Z., Zheng, L., Li, Y., Wang, S.: Linkage based face clustering via graph
  convolution network. In: CVPR (2019)

\bibitem{xiong2020layer}
Xiong, R., Yang, Y., He, D., Zheng, K., Zheng, S., Xing, C., Zhang, H., Lan,
  Y., Wang, L., Liu, T.: On layer normalization in the transformer
  architecture. In: ICML. pp. 10524--10533. PMLR (2020)

\bibitem{yang2020learning}
Yang, L., Chen, D., Zhan, X., Zhao, R., Loy, C.C., Lin, D.: Learning to cluster
  faces via confidence and connectivity estimation. In: CVPR (2020)

\bibitem{yang2019learning}
Yang, L., Zhan, X., Chen, D., Yan, J., Loy, C.C., Lin, D.: Learning to cluster
  faces on an affinity graph. In: CVPR (2019)

\bibitem{zhan2018consensus}
Zhan, X., Liu, Z., Yan, J., Lin, D., Loy, C.C.: Consensus-driven propagation in
  massive unlabeled data for face recognition. In: Proceedings of the European
  Conference on Computer Vision (ECCV). pp. 568--583 (2018)

\bibitem{zhao2003face}
Zhao, W., Chellappa, R., Phillips, P.J., Rosenfeld, A.: Face recognition: A
  literature survey. ACM computing surveys (CSUR)  \textbf{35}(4),  399--458
  (2003)

\end{thebibliography}
